\documentclass{article}

     \PassOptionsToPackage{numbers, compress}{natbib}

     \usepackage[final]{neurips_2019}

\usepackage{natbib}
\usepackage{ulem}
\usepackage{amsmath,amssymb,amsfonts}
\usepackage{algorithmic}
\usepackage{graphicx}
\usepackage{textcomp}
\usepackage{xcolor}
\usepackage{amsfonts}
\usepackage{amsthm}
\usepackage{algorithm}
\usepackage{amsmath}
\usepackage{bm}
\usepackage{mathtools}
\usepackage{multirow}
\DeclareMathOperator*{\argmax}{\arg\!\max}
\DeclareMathOperator*{\argmin}{\arg\!\min}

\newcommand{\zhaoqian}[1]{\textcolor{black}{{}#1}} 
\usepackage[utf8]{inputenc} 
\usepackage[T1]{fontenc}    
\usepackage{hyperref}       
\usepackage{url}            
\usepackage{booktabs}       
\usepackage{amsfonts}       
\usepackage{nicefrac}       
\usepackage{microtype}      

\usepackage{ulem}

\newif\ifshin
\shinfalse 
\newcommand\shin[2]{\ifshin{\sout{#1} \color{red}{#2}}\else{#2}\fi}

\title{Neural Networks Weights Quantization: Target None-retraining Ternary (TNT)}

\author{%
  Tianyu Zhang $^*$\\
  WeBank\\
  Shenzhen, Guangdong, China\\
  \texttt{brutuszhang@webank.com} \\
  \And
  Lei Zhu \thanks{Zhang and Zhu are the co-first authors, and Zhang is the corresponding author in this work. EMC2: 5th Edition Co-located with NIPS’19. Code: https://github.com/tenngre/TNT}\\
  Harbin Engineering University \\
  Harbin, Heilongjiang, China \\
  \texttt{zhulei@hrbeu.edu.cn } \\
  \AND
  Qian Zhao \\
  University of Hyogo \\
  Kobe, Hyogo, Japan\\
  \texttt{zhaoqian\_sunny@yahoo.co.jp} \\
  \And
  Kilho Shin \\
  Gakushuin University \\
  Tokyo, Japan \\
  \texttt{yoshihiro.shin@gakushuin.ac.jp} \\
}

\begin{document}

\maketitle

\begin{abstract}
  \shin{Quantifying weights (binary and ternary networks) of deep neural networks (DNN) is one solution for deploying inferences of DNNs on edge devices, such as mobiles, ASICs and FPGAs, because they have no sufficient resources to support the computation of millions of high precision weights and multiply-accumulate operations. This paper proposed a cosine similarity based target non-retraining ternary (TNT) compression method that reduces the searching range of a ternary weights from $3^{N}$ to $N$ and the computational complexity for the optimal ternary weights is only $O(N\log(N))$, moreover, it works on binary weights. Further, we found the upper limit of similarity between the target weights and ternary weights by TNT.}{Quantization of weights of deep neural networks (DNN) has proven to be an effective solution for the purpose of implementing DNNs on edge devices such as mobiles, ASICs and FPGAs, because they have no sufficient resources to support computation involving millions of high precision weights and multiply-accumulate operations. This paper proposes a novel method to compress vectors of high precision weights of DNNs to ternary vectors, namely a cosine similarity based target non-retraining ternary (TNT) compression method. Our method leverages cosine similarity instead of Euclidean distances as commonly used in the literature and succeeds in reducing the size of the search space to find optimal ternary vectors from $3^{N}$ to $N$, where $N$ is the dimension of target vectors. As a result, the computational complexity for TNT to find theoretically optimal ternary vectors is only $O(N \log(N))$. Moreover, our experiments show that, when we ternarize models of DNN with high precision parameters, the obtained quantized models can exhibit sufficiently high accuracy so that re-training models is not necessary.}
\end{abstract}

\section{Introduction}

\shin{Quantifying}{Quantizing} deep neural networks (DNNs) can reduce memory requirements and energy consumption when deploying inferences on edge devices, such as mobiles, ASICs and FPGAs. Comparing with \shin{other quantifying}{networks quantized by other} methods, \shin{}{the} binary and ternary networks use only $1$ or $2$ bits to represent DNNs' weights, \shin{that}{and therefore,} can further improve the performance of \shin{inference}{inferences} of DNN on edge devices because they not only remove multiplication operations but use less memory as well. As a result, many researches focus on binary and ternary quantifications.

BinaryConnect\cite{courbariaux2015binaryconnect} proposed a sign function to binarize the weights. Binary Weight Network (BWN) \cite{rastegari2016xnor} introduced the same binarization function but added an extra scaling factor to obtain better results. BinaryNet \cite{hubara2016binarized} and XNOR-Net \cite{rastegari2016xnor} extended the previous works \shin{}{so} that both weights and activations were binarized.
Instead of binarization, \shin{ternary networks}{ternarization}, which inherently \shin{prune}{prunes} \shin{the smaller weights}{weights close to zero} by setting them to zero during training\shin{, which makes network}{ to make networks} sparser\shin{}{,} \shin{are}{is} further studied.
TWN \cite{li2016ternary} quantized full precision weights to ternary weights \shin{by an approximated ternary function}{so} that the Euclidean distance (Second Normal Form)
between the full precision weights and the resulting ternary weights along with a scaling factor is minimized.
GXNOR-Net \cite{deng2018gxnor} provided a unified discretization framework for both weights and activations. Alemdar et al. \cite{alemdar2017ternary} trained ternary neural networks using a teacher-student approach based on a layer-wise greedy method. Mellempudi et al. \cite{mellempudi2017ternary} proposed a fine-grained quantization (FGQ) to ternarize pre-trained full precision models, while also constraining activations to 8 and 4 bits.

The \shin{most consumption parts in inference computation}
{parts in inference computation that consume time and energy in the largest scale}
involve many weights \shin{}{in computation,}
which are saved as tensors in every layer.
A tensor can be decomposed to a set of vectors, \shin{we call them}{referred to as} target vectors,
\shin{}{and each target vector is approximated to a binary or ternary vector.}
\shin{Therefore, }{To control the approximation error},
Euclidean distance is \shin{}{the most commonly} utilized
\shin{by many}{in many previous works in:}
\shin{quantification}{these quantization} methods \shin{to}{}measure
the \shin{}{approximation error or} similarity between original target vectors and
the approximated ternary or binary vectors \shin{}{as Euclidean distances}.
\shin{However, the computation is}{This method, however, is known to require} expensive \shin{}{computation.}
\shin{since the \shin{calculation}{computational complexity} is related to the dimension of \shin{the}{} target vectors. For example, the searching range of \shin{a}{an} $N$ dimensional vector is $3^{N}$, and its computational complexity to find a convincing result is $O(N^3)$ according to a previous work \cite{mellempudi2017ternary}. Therefore, a fast and reliable \shin{quantification}{quantization} method is needed.}
{For example, the time complexity of the tenary method proposed in \cite{mellempudi2017ternary} was $O(N^3)$.} In this paper, we \shin{proposed}{propose}
\shin{a cosine similarity based target non-retraining ternary (TNT) method, which reduce the searching range to $N$ and its computation complexity is only $O(N\log N)$. }
{a novel tenary method whose time complexity is improved to $O(N\log N)$ by replacing Euclidean distance by cosine similarity. We call our method a cosine similarity based target non-retraining ternary (TNT) method.}
\shin{Moreover, our work is nevertheless original in several regards}
{In addition, our method has following advantages}:
1) TNT is a non-retraining optimal \shin{quantification}{quantization} method for ternarization, binarization, and low bit-width quantizations;
2) We find the theoretical upper limit of similarity between target vectors and ternary vectors;
\shin{TNT guarantee to find}{it is guaranteed that TNT always finds} the optimal ternary vectors
\shin{}{with the maximum similarity} of original vectors;
3) We find the similarity is influenced by \shin{the distribution}{distributions of component values of}
\shin{the target vector's entries}{target vectors}, and \shin{}{furthermore,}
\shin{high}{higher} similarity can be obtained if \shin{the target vector's entries follow}{we assume}
\shin{an}{}uniform \shin{distribution}{distributions} \shin{which is better than the}{than} normal distributions.

\section{Method Description}

The proposed TNT first divides the tensor type weights of a DNN model into \shin{several}{plural} target vectors.
Then, it finds the ternary vector \shin{for}{most similar to} every target vector
\shin{by a}{with respect to} cosine similarity.
\shin{based technique which minimizes}{In other words, the ternary vector is selected so that} the intersection angle between the target vector and \shin{its ternary vectors}{ternary vector is minimized}.
Finally, it uses a scalar-tuning technique to adjust the error between one target vector and its ternary vector to obtain an optimal converting result.

\subsection{Tensor Decomposition and Vectorization}

The weights of a DNN are normally stored in a fourth-order tensor shape, such as $N \times C\times W \times H$, that contains $N$ third-order tensors and every third-order tensor has $C$ channels, $W$ width, and $H$ height. The purpose of tensor vectorization is to flatten every third-order tensor into \shin{several}{a set of} target vectors.
\shin{Decomposing}{We expect that decomposing} a tensor along the channel direction can yield good results,
\shin{due to our experiments for the reason that }
{because each channel is an integral unit which acts as a feature extractor for convolution calculations with a feature map. Hence, a third-order tensor can be vectorized to $\mathcal{W}^{(1)}, \mathcal{W}^{(2)}, \cdots, \mathcal{W}^{(C)}$. This expectation will be verified through experiments in this paper.}

\subsection{Target Non-retrain Ternarization}

We first introduce \shin{the}{our} cosine similarity \shin{technicals}{based technique} TNT,
which reduces the searching range to $N$. Then,
a scalar-tuning method is proposed to further optimize the ternary vector.
\shin{All}{The total} of the computational complexity is $O(N\log N)$.

\subsubsection{Cosine Similarity}

Given \shin{the}{a} target vector of a layer $j$ of a CNN,
which is \shin{}{denoted by} $\bm{w}^{(j)}=(w_{1}^{(j)}, \cdots, w_{N}^{(j)})$ \shin{,}{for}
$w_{i}^{(j)}\in \mathbb{R}$,
the purpose is to find a ternary vector $\bm{t}^{(j)}=(t_{1}^{(j)}, \cdots, t_{n}^{(j)}), t_{i}^{(j)} \in \{-1, 0, 1\}$
\shin{to replace it}{that approximates $\bm{w}^{(j)}$}.
For simple representation of equations, we eliminate the notation of $j$ since it only
\shin{represent}{represents} a layer $j$. 
In TNT, we use the cosine similarity metric between the two vectors
to find the \shin{}{optimal} ternary vector $\bm{t}$.
The cosine similarity between the target vector \shin{}{$\bm w$} and the ternary vector {$\bm t$} can be written as
Eq.(\ref{eq: cosine}),
where \shin{$\bm{\widehat{w}}$ is an unit vector}{$\bm{\widehat{w}} = \frac{\bm{w}}{\Vert\bm w\Vert_2}$
and $\alpha \in [0, \pi)$ is the intersecting angle between $\bm w$ and $\bm t$}.
The value of $\cos\alpha$ is controlled by vector $\bm{t}$ since every element $w_{i}$ in the target vector $\bm{w}$ is fixed.

\begin{eqnarray}
\label{eq: cosine}
\argmin_{\bm t}\alpha = \argmax_{\bm{t}}\frac{\bm{w}\cdot \bm{t}}{\|\bm{w}\|_{2} \|\bm{t}\|_{2}}
= \argmax_{\bm{t}}\frac{\bm{\widehat{w}} \cdot \bm{t}}{\|\bm{t}\|_{2}}
\end{eqnarray}

\shin{Since the unit vector can be written as }{When we denote} $\bm{\widehat{w}}=(a_{1}, a_{2}, \cdots, a_{N})$,
Eq.~(\ref{eq: cosine}) can be transformed to Eq.~(\ref{eq: cosine unit}), where $t_{i} \in \{-1, 0, 1\}$ and the search range of \shin{$t_{i}$}{$\bm t$} is $\{-1, 0, 1\}^{N}$.

\begin{eqnarray}
\label{eq: cosine unit}
  \argmax_{\bm{t}}\frac{\bm{\widehat{w}} \cdot \bm{t}}{\|\bm{t}\|_{2}}
  &=& \argmax_{t_{i}}\frac{\sum^{N}_{i=1}a_{i}t_{i}}{\sqrt{\sum^{N}_{i=1}(t_{i})^{2}}}
\end{eqnarray}

\zhaoqian{Let $(b_{1}, \dots, b_{N})$ be the propagation obtained by sorting $(\vert a_{1}\vert, \dots, \vert a_{N}\vert)$ in a decreasing order. \shin{A new sequence of $\left(t_{1}, t_{2}, \cdots, t_{N}\right)$ is generated at the same time according to the relation between $a_{i}$ and $t_{i}$. For the given $\bm{t}$ where the number of nonzero elements is $M$, i.e. $t_{i}$ where $i \in (M+1, \cdots, N)$ are all zeros. For an arbitrary choice of $t_{i}$ with $\sum_{i=1}^{N} \vert t_{i}\vert = M$, the following inequality holds:}{ Without loss of generality, we can assume that all $a_i$ are non-zero. First, we solve Eq.~(\ref{eq: cosine unit}) under the constraint of $\sum_{i=1}^{N} \vert t_{i}\vert = M$. Evidently,}}
\begin{eqnarray}
  \frac{\sum_{i=1}^N a_i t_i}{\sqrt M} \leq \frac{\sum_{j=1}^M b_j}{\sqrt M}.
\end{eqnarray}
\shin{\zhaoqian{Equation $(3)$ takes equal sign iff all $a_{i}$ and $t_{i}$ have the same sign}}
{holds, and the equality holds, if $t_i = 0$ for $a_i$ that corresponds to $b_j$ for $j \in \{M+1, \dots, N\}$
and $t_i = \frac{a_i}{\vert a_i\vert}$ for the others}.
Therefore, what we need to know is
\[
  \argmax\left\{\frac{\sum_{i=1}^M b_i}{\sqrt M}
      \Big\vert M = 1, \dots, N \right\},
\]
\shin{}{and} hence, calculating argument $\bm{t}$ in Eq.\ref{eq: cosine} equals to find the maximum value among $N$ candidates instead of among $3^N$ candidates. Moreover, the computational cost of finding $\bm{t}$ simply equals to \shin{calculating}{} the time complexity of sorting $\left(|a_{1}|, |a_{2}|, \cdots, |a_{N}|\right)$ to $\left(b_{1}, b_{2}, \cdots, b_{M}\right)$, which is $O(N\log N)$.

\subsubsection{Scalar-Tuning}

\begin{figure}[ht]
  \centering
  \includegraphics[width = 0.2\linewidth]{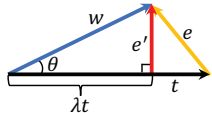}
  \caption{Scalar Constant $\lambda$ of a Ternary Vector}
  \label{fig:Scalar}
\end{figure}

\shin{Although near optimal ternary vectors can be found by minimizing the intersection angle between the target and the ternary vectors, which ensures they have similar directions. However, the error, that defined as $e = \|\bm{w} - \bm{t}\|$ in Fig. \ref{fig:Scalar}, between the target and ternary vectors can be still large yet.
To reduce the error $\Vert \bm{w} - \bm{t}\Vert_2$, it is efficient to memorize the length $\lambda$ of the orthographic projection of $w$ to $t$ in addition to $t$, since $\Vert \bm{w} - \lambda\bm{t}\Vert_2 \le \Vert \bm{w} - \bm{t}\Vert_2$ holds true (Fig. \ref{fig:Scalar}). The orthographic projection of $\bm{w}$ to $\bm{t}$ is
}
{Thus,
  we can obtain $\bm t$ whose intersecting angle with $\bm w$ is minimized.
  In other words,
  $\bm t$ approximately determines the direction of $\bm w$.
  To describe $\bm w$,
  we need to determine the length $\lambda$ in the direction of $\bm t$.
  The principle is to find an optimal $\lambda > 0$
  that minimizes the error $\left\Vert\bm w - \lambda \frac{\bm t}{\Vert\bm t\Vert_2}\right\Vert_2$.
  It is well known that the error is minimum, if, and only if,
  $\lambda \frac{\bm t}{\Vert\bm t\Vert_2}$ is the orthographic projection of $\bm{w}$ to $\bm{t}$ (Fig. \ref{fig:Scalar}),
  which is given by
}
\[
  \lambda \frac{\bm{t}}{\Vert \bm{t}\Vert_2} = \frac{\bm{w} \cdot \bm{t}}{\bm{t} \cdot \bm{t}} \bm{t}.
\]

This increases the necessary memory size (footprint), but is effective to improve accuracy.
Moreover, if $\bm{t}$ includes both positive and negative elements, we could improve the accuracy more by memorizing one more scalar: We let $\bm{t} = \bm{t}_p - \bm{t}_n$ with $\bm{t}_p \ge 0$ and $\bm{t}_n \le 0$ and
$\bm{w} = \bm{w}_p - \bm{w}_n$ with $\bm{w}_p \ge 0$ and $\bm{w}_n \le 0$:
for a vector $\bm{v}$,
$\bm{v} \ge 0$ ($\bm{v} \le 0$) means that all the elements of $\bm{v}$ is non-negative (non-positive).
We should note that
$\bm{t}_p\cdot \bm{t}_n = \bm{t}_p\cdot \bm{w}_n = \bm{w}_p \cdot \bm{t}_n = \bm{w}_p \cdot \bm{w}_n = 0$, $\bm{t}_p\cdot \bm{w}_p \ge 0$ and $\bm{t}_n\cdot \bm{w}_n \ge 0$ holds.
Therefore, we have:

\begin{align*}
  \left\Vert \bm{w} - \frac{\bm{w} \cdot \bm{t}}{\bm{t} \cdot \bm{t}} \bm{t} \right\Vert_2^2
  &
    = \bm{w} \cdot \bm{w} - \frac{(\bm{w}\cdot \bm{t})^2}{\bm{t}\cdot \bm{t}}
    \\&
     = \bm{w}_p\cdot \bm{w}_p + \bm{w}_n\cdot \bm{w}_n - \frac{(\bm{w}_p\cdot \bm{t}_p + \bm{w}_n\cdot \bm{t}_n)^2}{\bm{t}_p\cdot \bm{t}_p + \bm{t}_n\cdot \bm{t}_n}
 \\&
  \ge \bm{w}_p\cdot \bm{w}_p - \frac{(\bm{w}_p\cdot \bm{t}_p)^2}{\bm{t}_p\cdot \bm{t}_p}
  + \bm{w}_n\cdot \bm{w}_n - \frac{(\bm{w}_n\cdot \bm{t}_n)^2}{\bm{t}_n\cdot \bm{t}_n}
  \\&
  = \left\Vert \bm{w}_p - \frac{\bm{w}_p \cdot \bm{t}_p}{\bm{t}_p \cdot \bm{t}_p} \bm{t}_p \right\Vert_2^2
  + \left\Vert \bm{w}_n - \frac{\bm{w}_n \cdot \bm{t}_n}{\bm{t}_n \cdot \bm{t}_n} \bm{t}_n \right\Vert_2^2.
\end{align*}

Thus, if we let $\lambda_p = \frac{\bm{w}_p \cdot \bm{t}_p}{\Vert \bm{t}_p \Vert}$ and
$\lambda_n = \frac{\bm{w}_n \cdot \bm{t}_n}{\Vert \bm{t}_n \Vert}$ memorized in addition to $\bm{t}$,
we can not only save memory size but also suppress loss of accuracy.

\section{Simulations}

In this part, we first show the performance of our TNT method on transforming target vectors to ternary vectors. Then, we show the upper limit of similarity of ternary and binary when utilizing different distributions to initialize target vectors. Finally, we demonstrate an example using TNT to convert weights of DNN models to ternary. All experiments are run on a PC with Intel(R) Core(TM) i7-8700 CPU at 3.2GHz using 32GB of RAM and a NVIDIA GeForce GTX 1080 graphics card, running Windows 10 system.

\subsection{Converting Performance}

\begin{figure}[ht]
\centering{
\includegraphics[width = 1\linewidth]{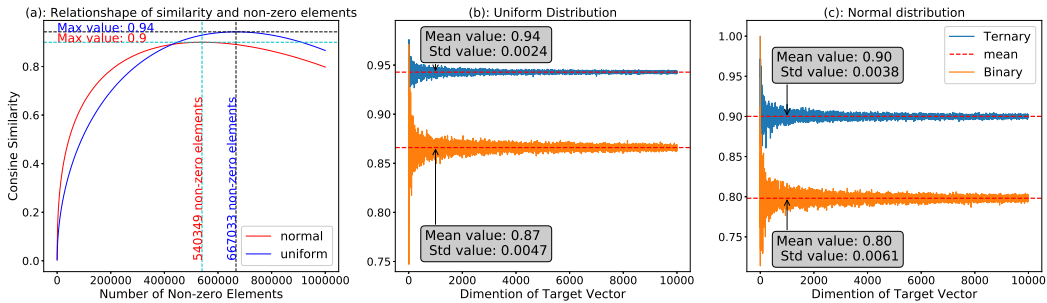}}
\caption{Simulation Result of Cosine Similarity by TNT method}
\label{fig:Uniform-norm}
\end{figure}

\shin{We evaluate the converting performance by observing the cosine similarity between a target vector and its corresponding ternary vector. Figure \ref{fig:Uniform-norm} (\textsl{a}) shows the cosine similarity simulation result where the dimension of a ternary vector is $1000000$.}
{In order to investigate the accuracy performance of our ternarization method,
  we prepare two target vectors of dimension 1,000,000:
  one has elements determined independently following a uniform distribution;
  The other follows a normal distribution instead.
  Figure \ref{fig:Uniform-norm} (\textsl{a}) shows the cosine similarity scores observed
  when we change the number $M$ of non-zero elements of $\bm t$.}
\shin{We can see that if the entries of the target vector follow a uniform distribution (blue curve), the maximum cosine similarity is higher than \shin{that}{when} the entries of the target vector follow a normal distribution (red curve), and the cosine similarities are \shin{$94.28\%$}{$0.94$} and \shin{$90.00\%$}{$0.90$}, respectively.
On the other hand, we can find that the ternary vector followed uniform distribution has more non-zero entries, which is $667033$, than the ternary vector under normal distribution, which is $540349$, at the maximum cosine similarity point. }
{The highest score for the target vector that has followed a uniform distribution is $0.94$
  when $667,033$ elements of $\bm t$ are non-zero,
  while the highest score is $0.90$ for a normal distribution when $540,349$ elements of $\bm t$ are non-zero.}
\shin{Moreover, the curve of cosine similarity based TNT is a unimodal cure which has the maximum value, which means the TNT can provide a criterion to evaluate the upper limit of the best quantization results.}
{The curves of the cosine similarity score are unimodal,
  and if this always holds true,
  finding maximum cosine similarity scores can be easier.}

Moreover, we found a fact that the cosine similarity is not easily affected by the dimension of a ternary or binary vector. We calculated 10000 times of the maximum cosine similarity with the dimension of target vectors increases
\shin{one}{by one} at each time. Figure \ref{fig:Uniform-norm} (\textsl{b}) and (\textsl{c}) show the simulation results: 1) regardless of the target vector under normal distribution or uniform distribution, ternary vectors reserve a higher similarity. 2) \shin{he}{the} cosine similarity of ternary and binary vectors converge to a stable value with the increasing of vector dimension, and the ternary vector has a smaller variance comparing with the binary vector.

\subsection{Performance on Neural Networks}

We perform our experiments on LeNet-5\cite{li2016ternary}, VGG-7\cite{li2016ternary}, and VGG16\cite{simonyan2014very} using MNIST, CIFAR-10, and ImageNet datasets respectively to first train a full precision network model, and then replace the floating point \shin{parameter}{parameters} by the ternary \shin{parameter}{parameters} obtained by the proposed TNT. A precise comparison between floating point model and ternary model \shin{are}{is} conducted.

The experiment results are shown in Table \ref{TNTPerformance}. It shows that, without network retraining, inferences with ternary parameters only lose $0.21\%$ and $0.14\%$ of accuracies using LeNet-5 and VGG-7 respectively on MNIST dataset. And it loses $2.22\%$ of accuracy for VGG-7 on CIFAR-10 dataset. For VGG-16 network on ImageNet dataset, the Top-1 and Top-5 accuracy dropped $8\%$ and $5.34\%$, respectively. Moreover, the memory size of LeNet-5 and VGG-7 are reduced $16$ times since each ternary weight only requires $2$ bits of memory. On the other hand, because of converting the first and last layer of VGG-16 to ternary without fine-tuning has a significant affection on the accuracy, which is the same phenomenon mentioned in \cite{mellempudi2017ternary}, we do not convert the first and the last layer in VGG-16, and the parameter size reduces $11.1$ times.


\begin{table}[]
\small
\centering
\caption{TNT Performance on Neural Networks}
\label{TNTPerformance}
\begin{tabular}{|c|c|c|c|c|}
\hline
Network                                                     & Base Line        & TNT Converted                           & Parameters & Converting Times \\ \hline
\begin{tabular}[c]{@{}c@{}}LeNet-5\\ (MNIST)\end{tabular}   & 99.18\%          & {\color[HTML]{FE0000} 98.97\%}          & 1,663,370  & 7.803s           \\ \hline
\begin{tabular}[c]{@{}c@{}}VGG-7 \\ (CIFAR-10)\end{tabular} & 91.31\%          & {\color[HTML]{FE0000} 89.09\%}          & 7,734,410  & 88.288s          \\ \hline
\begin{tabular}[c]{@{}c@{}}VGG-16\\ (ImageNet)\end{tabular} & 64.26\%, 85.59\% & {\color[HTML]{FE0000} 56.26\%, 80.25\%} & 12,976,266 & 115.863s         \\ \hline
\end{tabular}
\end{table}

\section{Conclusions}

\zhaoqian{In this paper, we proposed a target non-retraining ternary (TNT) method to convert a full precision parameters model to a ternary parameters model accurately and quickly without retraining of the network. In our approach, firstly, we \shin{reduced}{succeeded in reducing} \shin{}{the size of} the searching range from $3^{N}$ to $N$ by evaluating the \shin{}{cosine} similarity between \shin{}{a} target vector and \shin{}{a} ternary vector. \shin{by cosine similarity}{} Secondly, scaling-tuning factors are proposed coupling with the cosine similarity to further enable the TNT to find the best ternary vector.
Due to the smart tricks, TNT's computational complexity \shin{only cost}{is only} $O(N\log N)$.
Thirdly, we \shin{identify}{showed} that the distributions of parameters \shin{has}{have} an obvious affection on the weight converting result. \shin{which gives us a hint to pay attention to initialize parameters, if we want to convert a model to ternary or binary type}{This implies that the initial distributions for parameters are important}.
Moreover, we \shin{tested}{applied} the TNT \shin{on}{to} several models. \shin{that identify the it can convert a model with acceptable accuracy loss.}
{As a result, we verified that quantization by our TNT method caused a small loss of accuracy.}}

\bibliography{egbib}
\bibliographystyle{IEEEtran}
\end{document}